\documentclass{bmvc2k}

\usepackage{times}
\usepackage{epsfig}
\usepackage{graphicx}
\usepackage{amsmath}
\usepackage{amssymb}
\usepackage[misc]{ifsym}

\usepackage{booktabs}
\usepackage[x11names,dvipsnames,table]{xcolor} 
\usepackage{multirow}
\usepackage{array}
\newcolumntype{L}[1]{>{\raggedright\let\newline\\\arraybackslash\hspace{0pt}}m{#1}}
\newcolumntype{C}[1]{>{\centering\let\newline\\\arraybackslash\hspace{0pt}}m{#1}}
\newcolumntype{R}[1]{>{\raggedleft\let\newline\\\arraybackslash\hspace{0pt}}m{#1}}

\title{Detecting Semantic Parts on Partially Occluded Objects}

\addauthor{Jianyu Wang*}{wjyouch@gmail.com}{1}
\addauthor{Cihang Xie*}{cihangxie306@gmail.com}{2}
\addauthor{Zhishuai Zhang*}{zhshuai.zhang@gmail.com}{2}
\addauthor{Jun Zhu}{zhujun.sjtu@gmail.com}{1}
\addauthor{Lingxi Xie$^{(\textrm{\Letter})}$}{198808xc@gmail.com}{2}
\addauthor{Alan L. Yuille}{alan.l.yuille@gmail.com}{2}

\addinstitution{
Baidu Research (USA), \\
Sunnyvale, CA 94089 USA
}
\addinstitution{
Department of Computer Science, \\
The Johns Hopkins University, \\
Baltimore, MD 21218 USA \\
{\textcolor{white}{.}} \\
{\textcolor{white}{.}} \\
{\textcolor{white}{.}} \\
{\textcolor{white}{.}} \\
{\small * This work was done when the first author was a Ph.D. student at UCLA.
The first three authors contributed equally to this work.} \\
}

\runninghead{Wang et.al.}{Detecting Semantic Parts on Partially Occluded Objects}

\begin{document}

\maketitle

\begin{abstract}
In this paper, we address the task of detecting semantic parts on partially occluded objects.
We consider a scenario where the model is trained using non-occluded images but tested on occluded images.
The motivation is that there are infinite number of occlusion patterns in real world,
which cannot be fully covered in the training data.
So the models should be inherently robust and adaptive to occlusions
instead of fitting / learning the occlusion patterns in the training data.
Our approach detects semantic parts by accumulating the confidence of local visual cues.
Specifically, the method uses a simple voting method, based on log-likelihood ratio tests and spatial constraints,
to combine the evidence of local cues.
These cues are called {\em visual concepts}, which are derived by clustering the internal states of deep networks.
We evaluate our voting scheme on the VehicleSemanticPart dataset with dense part annotations.
We randomly place two, three or four irrelevant objects onto the target object to generate testing images with various occlusions.
Experiments show that our algorithm outperforms several competitors in semantic part detection when occlusions are present.
\end{abstract}

\section{Introduction}
\label{Introduction}

\noindent
{\em ``The absence of evidence is not the evidence of absence.''}\hfill{\textsc{ --- A Proverb}}

Deep neural networks~\cite{Krizhevsky_2012_ImageNet} have been successful on a wide range of vision tasks
and in particular on object detection~\cite{Girshick_2014_Rich}\cite{Ren_2015_Faster}.
There have, however, been much fewer studies on semantic part detection.
Here, a {\em semantic part} refers to a fraction of an object which can be verbally described,
like a wheel of a car or a wing of an airplane.
Detecting semantic parts of objects is a very important task, which enables us to parse the object and reason about its properties.
More importantly, humans are able to recognize occluded objects by looking at parts,
{\em e.g.}, simply seeing a {\em car wheel} is often enough to infer the presence of the entire {\em car}.

In this paper we study the problem of detecting the semantic parts of partially occluded objects.
There are in general two strategies for occlusion handling:
making the model inherently robust to occlusions or fitting the model to occlusion patterns in the training set.
We argue that the latter strategy is inferior because there are infinite number of occlusion patterns in real world,
so the training set only contains a biased subset of occlusion patterns.
Therefore, we consider the scenario, where the part detector is trained using non-occluded images but tested on occluded images.
In other words, the distribution of testing images are very different from that of training images.
This problem setting favors learning the models which are naturally robust and adaptive to occlusions
instead of over-fitting the occlusion patterns in the training data.

\begin{figure}[t!]
\centering
    \includegraphics[width=\textwidth]{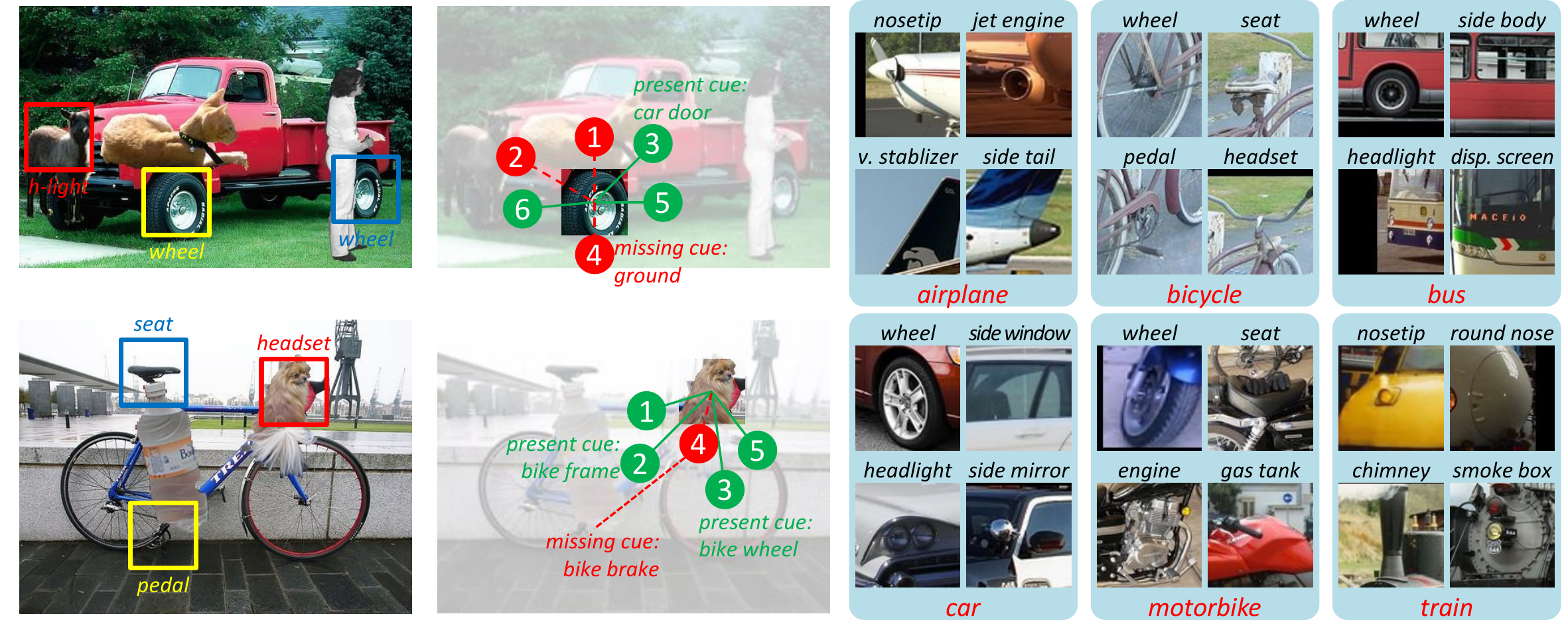}
\caption{
    Left: our goal is to detect semantic parts, in particular under occlusion.
    Red, blue and yellow boxes indicate fully-, partially- and non-occluded semantic parts respectively.
    The proposed voting method is able to switch on/off visual cues (green/red disks) for semantic parts detection.
    Right: typical semantic parts on six types of rigid objects from VehicleSemanticPart dataset~\cite{Wang_2015_Unsupervised}.
    Some semantic parts ({\em e.g.}, {\em wheel}) can appear in different classes,
    while some others ({\em e.g.}, {\em chimney}) only appear in one class.
    This figure is best viewed in color.
}
\label{Fig:Goal}
\end{figure}

Figure~\ref{Fig:Goal} illustrates the task we are going to address and some typical examples during testing.
Since some of the target semantic parts are partially occluded,
the state-of-the-art holistic object detectors such as Faster-RCNN~\cite{Ren_2015_Faster}
are sometimes unable to provide satisfying results due to their lack of ability to deal with occlusions.
When the testing image contains an occlusion pattern which does not appear in the training set,
such detectors can fail in finding a proper object proposal and/or making classification based on the detected region-of-interest.
For example, if a {\em car wheel} is occluded by a large {\em table},
there might be no proposal covering the {\em wheel} (low recall in objectness detection),
and the classifier may also be confused even if a perfect proposal is given (low accuracy in object recognition).
This inspires us to detect parts by accumulating local visual cues,
instead of directly learning a holistic template as what Faster-RCNN~\cite{Ren_2015_Faster} is essentially doing.

We start with the recent work~\cite{Wang_2015_Unsupervised} which showed that deep networks have internal representations,
which are called {\em visual concepts}.
They are related to semantic parts and can be used for part detection.
In this paper, we show that visual concepts can be combined together to detect semantic parts.
We design a novel voting scheme, which is built upon some simple techniques such as log-likelihood ratio tests and spatial pooling.
In the training phase on non-occluded objects,
we find the relationship between each semantic part and its supporting visual concepts,
and take into account their relative positions to model the spatial contexts.
In the testing phase on occluded objects,
these clues are integrated together to detect partially or even fully occluded semantic parts.
As shown in Figure~\ref{Fig:Goal}, our voting algorithm is adaptive to different contexts,
since it enjoys the flexibility of switching on/off visual cues and avoids using negative cues.

We evaluate our algorithm on VehicleSemanticPart dataset~\cite{Wang_2015_Unsupervised} (see Figure~\ref{Fig:Goal}),
which provides dense labeling of more than $100$ semantic parts over $6$ object classes.
In order to create the test set with various occlusion patterns,
we randomly superimpose two, three or four irrelevant objects (named occluders) onto the target object.
These occluders are manually labeled object segments
from the PASCAL-Parts dataset~\cite{Chen_2014_Detect}.
We also control the occlusion ratio by computing the fraction of occluded pixels on the target object.
Experiments reveal the advantage over several competitors in detection accuracy (measured by mean AP)
under the scenario where the target object is partially occluded.
Our approach, while being able to deal with occlusions, does not need to be trained on occluded images.

This paper is organized as follows.
Section~\ref{RelatedWork} discusses related work.
Our voting algorithm is described in Section~\ref{Algorithm}.
Section~\ref{Experiments} describes the experiments which validate our approach, and Section~\ref{Conclusions} concludes this work.

\section{Related Work}
\label{RelatedWork}

Object detection is a fundamental task in computer vision.
As the fast development of deep neural networks~\cite{Krizhevsky_2012_ImageNet}\cite{Simonyan_2015_Very}\cite{He_2016_Deep},
this field has been recently dominated by one type of pipeline~\cite{Girshick_2014_Rich}\cite{Ren_2015_Faster},
which first generates a set of object proposals~\cite{Alexe_2012_Measuring}\cite{Uijlings_2013_Selective},
and then predicts the object class of each proposal.
This framework has significantly outperformed conventional approaches,
which are based on handcrafted features~\cite{Dalal_2005_Histograms}
and deformable part models~\cite{Felzenszwalb_2010_Object}\cite{Azizpour_2012_Object}.

Visual concepts~\cite{Wang_2015_Unsupervised} are obtained by clustering the intermediate neural responses of deep networks.
It is shown that on rigid objects, image patches corresponding to the same visual concept are often visually similar,
and that visual concepts are fairly effective in detecting keypoints in the PASCAL3D+ dataset~\cite{Xiang_2014_Beyond}.
Our studies are built on previous studies which showed that
filters in deep networks often exhibited preferences in stimuli~\cite{Zhou_2015_Object}.

There are some works about detecting parts using deep networks.
In~\cite{Chen_2014_Articulated}, they used deep features as unary term and built graphical models to assemble parts into entire human.
\cite{Zhang_2014_Part} applied R-CNN~\cite{Girshick_2014_Rich} to detecting parts,
which were later used for fine-grained categorization.
\cite{Long_2014_Convnets} used deep feature with SVM for keypoint detection.
By contrast, some works used CNN features to discover mid-level visual elements in an unsupervised way,
which are then used for object / scene classification~\cite{Li_2015_Mid}\cite{Simon_2015_Neural}\cite{Xiao_2015_Application}.

Occlusion is a common difficulty in object detection~\cite{Kar_2015_Amodal} or segmentation~\cite{Li_2016_Amodal}.
\cite{Li_2014_Integrating} used And-or-Graph (AOG) to model the occlusion patterns for car detection.
Part-based models is very useful for detecting partially occluded
objects~\cite{Chen_2015_Parsing}\cite{Li_2014_Integrating}.
Our method is also part-based but applied to detecting semantic parts.

Our voting method is similar to~\cite{Leibe_2004_Combined}\cite{Maji_2009_Object}\cite{Okada_2009_Discriminative}.
But, we incorporate log-likelihood ratio test~\cite{Amit_2002_2D} and spatial constraint~\cite{Grimson_1990_Object}
as key components into the voting method, which is new.
Also we address a new problem of detecting semantic parts under occlusions,
where the training images and testing images are quite different.

\section{Our Algorithm}
\label{Algorithm}

\subsection{Notations}
\label{Algorithm:Notations}

We first introduce some notations.
Each semantic part $\mathrm{SP}_s$ has an index ${s}\in{\left\{1,2,\ldots,\left|\mathcal{S}\right|\right\}}$,
where $\mathcal{S}$ is a pre-defined set of all semantic parts.
Let $q$ denote a position at the input image lattice, {\em i.e.}, ${q}\in{\mathcal{L}_0}$.
Denote a position at the {\em pool-4} layer as ${p}\in{\mathcal{L}_4}$,
then a feature vector can be written as ${\mathbf{f}\!\left(\mathbf{I}_p\right)}\in{\mathbb{R}^{512}}$ ({\em e.g.}, VGG-16).
Most often, we need to consider the relationship between two positions
on two layers ${p}\in{\mathcal{L}_4}$ and ${q}\in{\mathcal{L}_0}$.
Let $\mathcal{L}_0\left(p\right)$ denote the exact mapping from $\mathcal{L}_4$ to $\mathcal{L}_0$.
Inversely, let ${\mathcal{L}_4\!\left(q\right)}={\arg\min_p\left\{\mathrm{Dist}\left(q,\mathcal{L}_0\left(p\right)\right)\right\}}$
denote the closest position at the $\mathcal{L}_4$ layer grid that corresponds to $q$.
We denote the neighborhood of ${q}\in{\mathcal{L}_0}$ on the $\mathcal{L}_4$ layer as
${\mathcal{N}\!\left(q\right)} \subset \mathcal{L}_4$,
which is defined as ${\mathcal{N}\!\left(q\right)}=
    {\left\{p\in\mathcal{L}_4\mid\mathrm{Dist}\left(q,\mathcal{L}_0\left(p\right)\right)<\mathrm{\gamma_{\mathrm{th}}}\right\}}$.
The neighborhood threshold $\gamma_{\mathrm{th}}$ is set to be $120$ pixels
and will be discussed later in Section~\ref{Experiments:NoOcclusion}.

Following~\cite{Wang_2015_Unsupervised}, we extract {\em pool-4} layer features using VGG-16~\cite{Simonyan_2015_Very}
from a set of training images,
and use $K$-Means to cluster them into a set $\mathcal{V}$ of visual concepts,
which corresponds to certain types of visual cues that appear on the image.
Here, each visual concept $\mathrm{VC}_v$ has an index ${v}\in{\left\{1,2,\ldots,\left|\mathcal{V}\right|\right\}}$.
The $v$-th clustering center is denoted as ${\mathbf{f}_v}\in{\mathbb{R}^{512}}$.

\begin{figure}[t]
\centering
    \includegraphics[width=\textwidth]{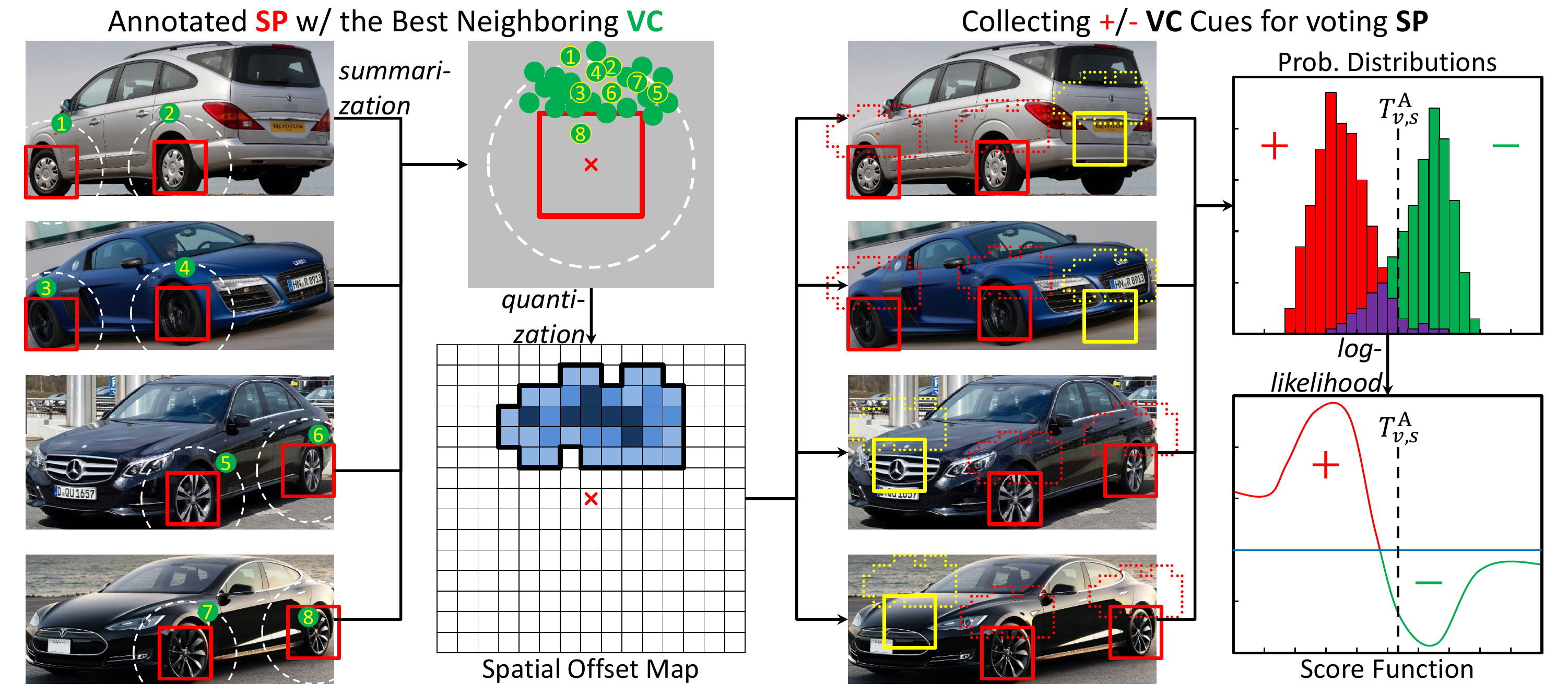}
\caption{
    Illustration of the training phase of one $\left(\mathrm{VC}_v,\mathrm{SP}_s\right)$ pair.
    Only one negative point is shown in each image (marked in a yellow frame in the third column),
    though the negative set is several times larger than the positive set.
    Each green dot indicates an offset $\Delta p^\star$ (see Section~\ref{Algorithm:Training:SpatialRelationship}),
    and the spatial offset map contains the frequencies of these offsets.
    The automatic determination of negative points,
    the probability distributions and the score functions are described in Section~\ref{Algorithm:Training:Evidences}.
}
\label{Fig:Illustration}
\end{figure}

Our algorithm is composed of a training phase and a testing phase, detailed in the following subsections.
The training phase is illustrated in Figure~\ref{Fig:Illustration}.
We perform training and testing on each semantic part individually.

\subsection{The Training Phase}
\label{Algorithm:Training}

The training phase starts with cropping each object according to the ground-truth object bounding box,
and rescaling it so that the short edge contains $224$ pixels.

\subsubsection{Spatial Relationship between Visual Concepts and Semantic Parts}
\label{Algorithm:Training:SpatialRelationship}

Our goal is to use visual concepts to detect semantic parts.
Therefore, it is important to model the spatial relationship of each $\left(\mathrm{VC}_v,\mathrm{SP}_s\right)$ pair.
Intuitively, a semantic part can be located via its neighboring visual concepts.
If a visual concept $\mathrm{VC}_v$ serves as a good visual cue to locate a semantic part $\mathrm{SP}_s$,
then the $\mathrm{SP}_s$ may only appear at some specified positions relative to $\mathrm{VC}_v$.
For example, if $\mathrm{VC}_v$ represents the upper part of a {\em wheel},
we shall expect the semantic part ({\em wheel}) to appear slightly below the position of $\mathrm{VC}_v$.
Motivated by this, we define a {\em spatial offset map} for each $\left(\mathrm{VC}_v,\mathrm{SP}_s\right)$ pair.
An offset map $\mathcal{H}_{v,s}$ is a set of frequencies of offsets $\Delta p$,
indicating the most likely spatial relationship between $\mathrm{VC}_v$ and $\mathrm{SP}_s$,
{\em i.e.}, if $\mathrm{VC}_v$ at position ${p}\in{\mathcal{L}_4}$ supports $\mathrm{SP}_s$,
$\mathrm{SP}_s$ may appear around $\mathcal{L}_0\!\left(p+\Delta p\right)$ in the image lattice.
Since the neighborhood threshold $\gamma_{\mathrm{th}}$ is $120$ pixels and the spatial stride of $\mathcal{L}_4$ is $16$ pixels,
$\mathcal{H}_{v,s}$ is a subset of $\left\{-7,-6,\ldots,7\right\}\times\left\{-7,-6,\ldots,7\right\}$,
or equivalently, a $15 \times 15$ grid.

To estimate the offset map $\mathcal{H}_{v,s}$,
we perform spatial statistics for each $\left(\mathrm{VC}_v,\mathrm{SP}_s\right)$ pair.
We find each annotated ground-truth position $q$,
and compute the position $p^\star$ in its $\mathcal{L}_4$ neighborhood which best stimulate $\mathrm{VC}_v$, {\em i.e.},
${p^\star}={{\arg\min_{p\in\mathcal{N}\!\left(q\right)}}\left\|\mathbf{f}\!\left(\mathbf{I}_p\right)-\mathbf{f}_v\right\|}$.
Then ${\Delta p^\star}={\mathcal{L}_4\!\left(q\right)-p^\star}$ is added to a score table.
After all annotated semantic parts are considered,
each offset in the score table gets a frequency ${\mathrm{Fr}\!\left(\Delta p\right)}\in{\left[0,1\right]}$.
The offsets with above-average frequencies compose the offset map $\mathcal{H}_{v,s}$.
We rewrite a neighborhood $\mathcal{N}\!\left(q\right)$ equipped with the offset map $\left(\mathrm{VC}_v,\mathrm{SP}_s\right)$
as $\mathcal{N}_{v,s}\!\left(q\right)$,
which contains all positions $\left\{p+\Delta p\in\mathcal{N}\!\left(q\right)\mid\Delta p\in\mathcal{H}_{v,s}\right\}$.

Some typical offset maps are shown in Figure~\ref{Fig:Training}.
We can see that the concentration ratio of an offset map can reflect, at least to some extent,
whether a visual concept is good for supporting or detecting the specified semantic part.
In the next step, we shall integrate these spatial cues to obtain a score function.

\subsubsection{Probabilistic Distributions, Supporting Visual Concepts and Log-likelihoods}
\label{Algorithm:Training:Evidences}

We quantify the evidence that a visual concept ${\mathrm{VC}_v}\in{\mathcal{V}}$ gives
for detecting a semantic part ${\mathrm{SP}_s}\in{\mathcal{S}}$ and also its ability to localize $\mathrm{SP}_s$.
We study this for all possible pairs $\left(\mathrm{VC}_v,\mathrm{SP}_s\right)$.
We find, not surprisingly, that a subset of visual concepts are helpful for detecting a semantic part while others are not.
We judge the quality of $\mathrm{VC}_v$ in detecting $\mathrm{SP}_s$
by measuring its ability to distinguish positive and negative visual cues.
This is done by estimating the distribution of Euclidean distance between $\mathrm{VC}_v$ and $\mathrm{SP}_s$.

\begin{figure*}
\centering
    \includegraphics[width=1.2in]{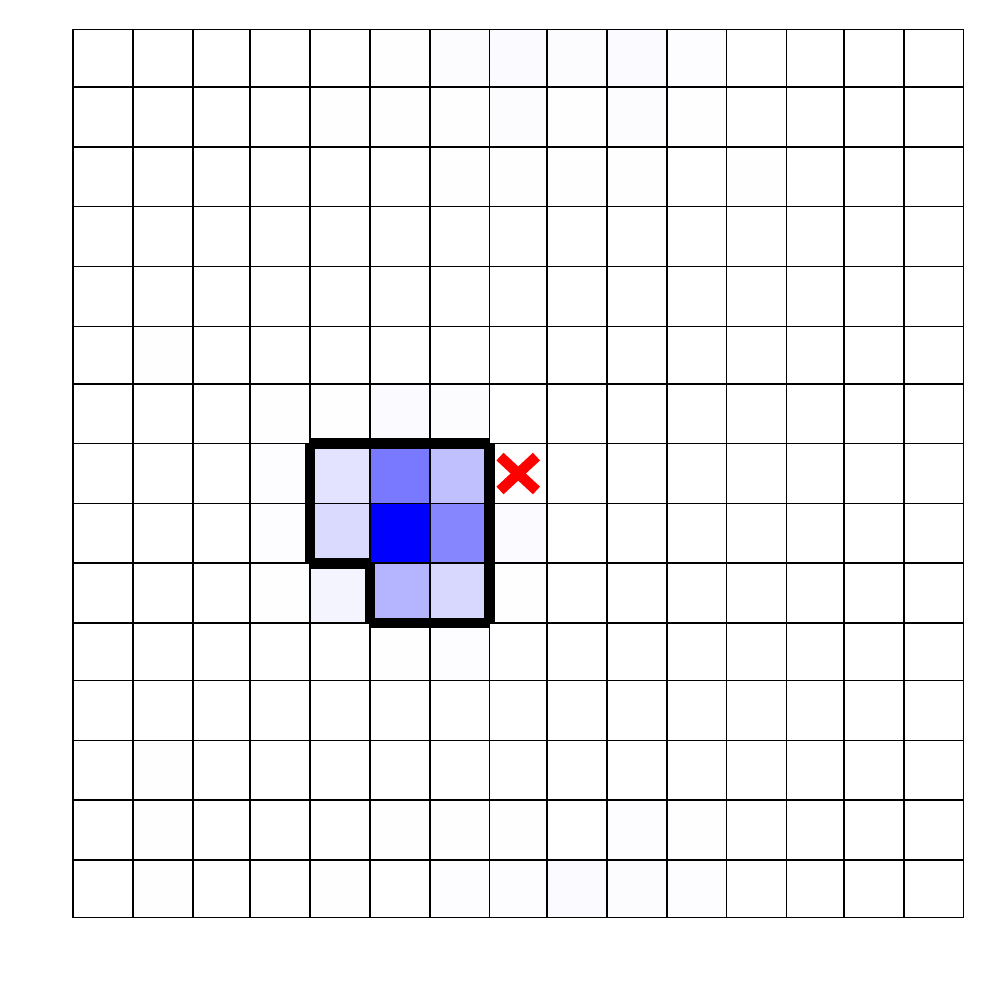}
    \includegraphics[width=1.2in]{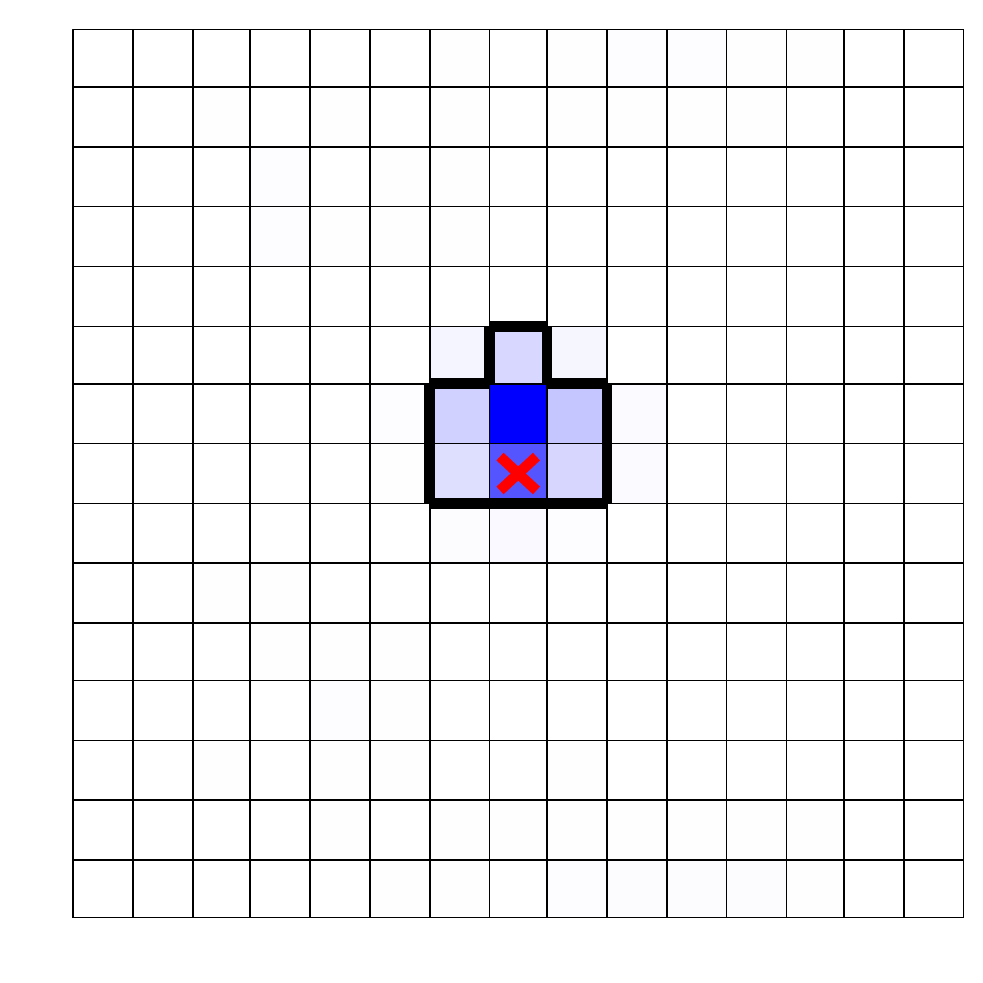}
    \includegraphics[width=1.2in]{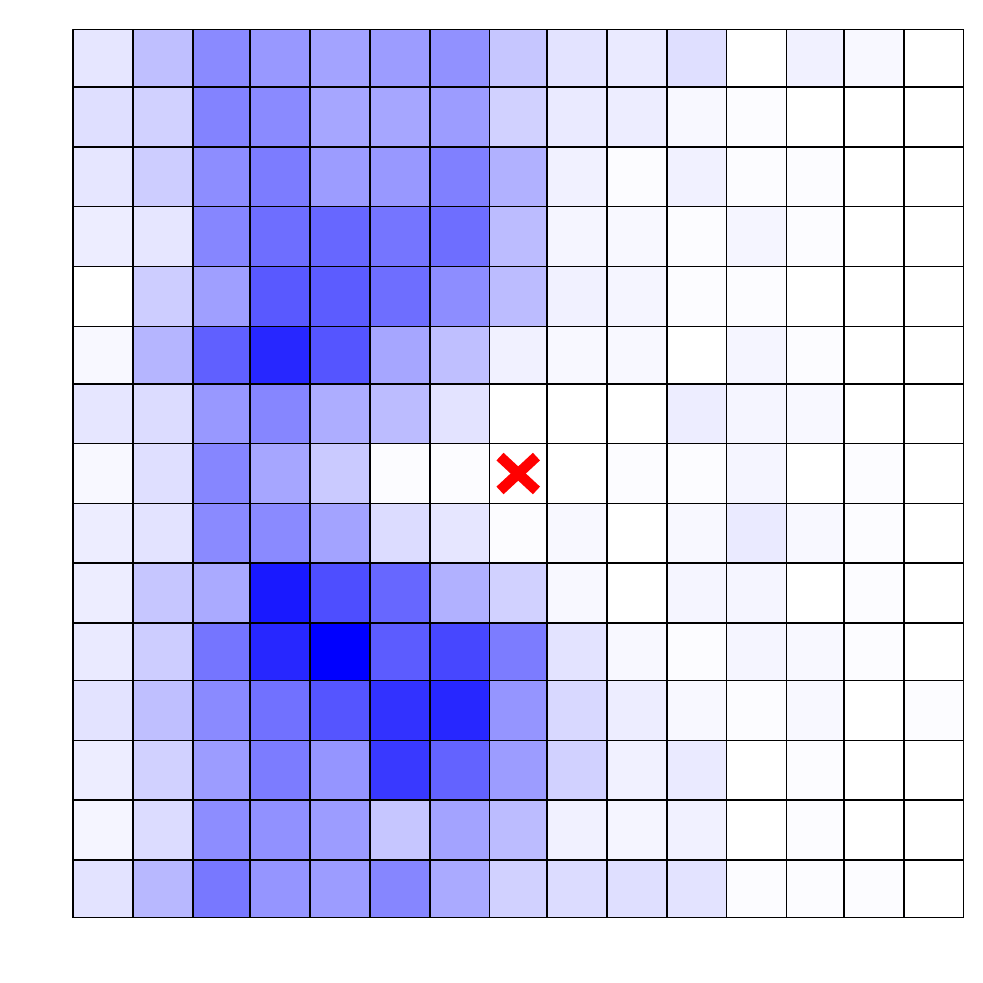}
    \includegraphics[width=1.2in]{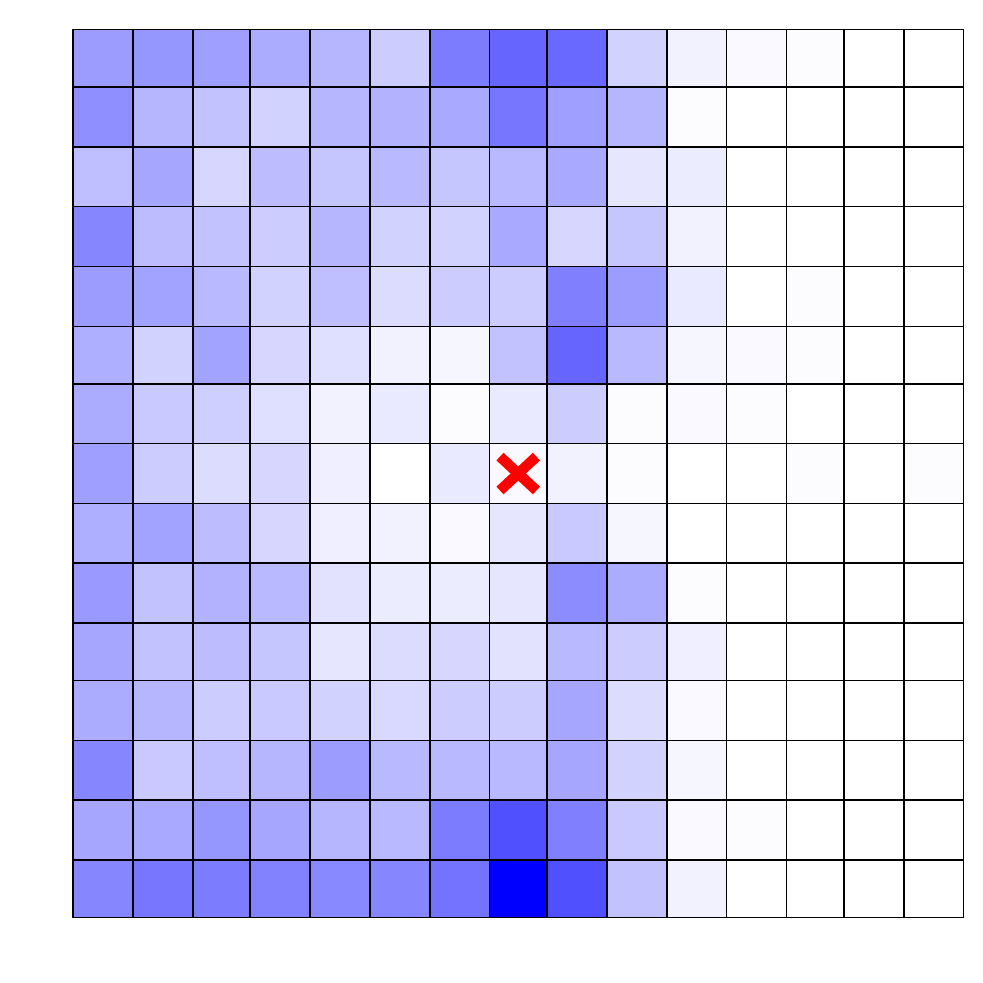}
    \includegraphics[width=1.2in]{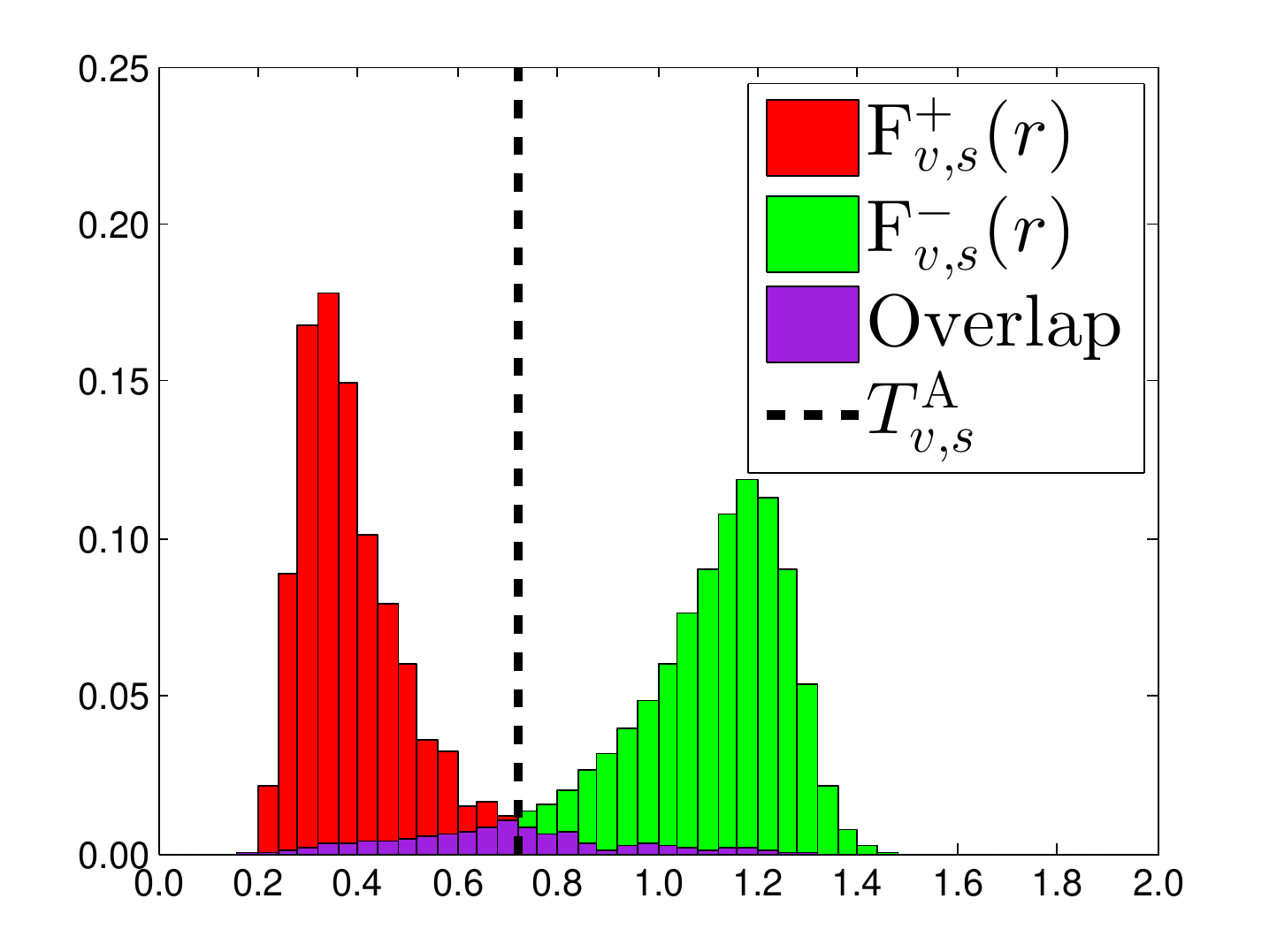}
    \includegraphics[width=1.2in]{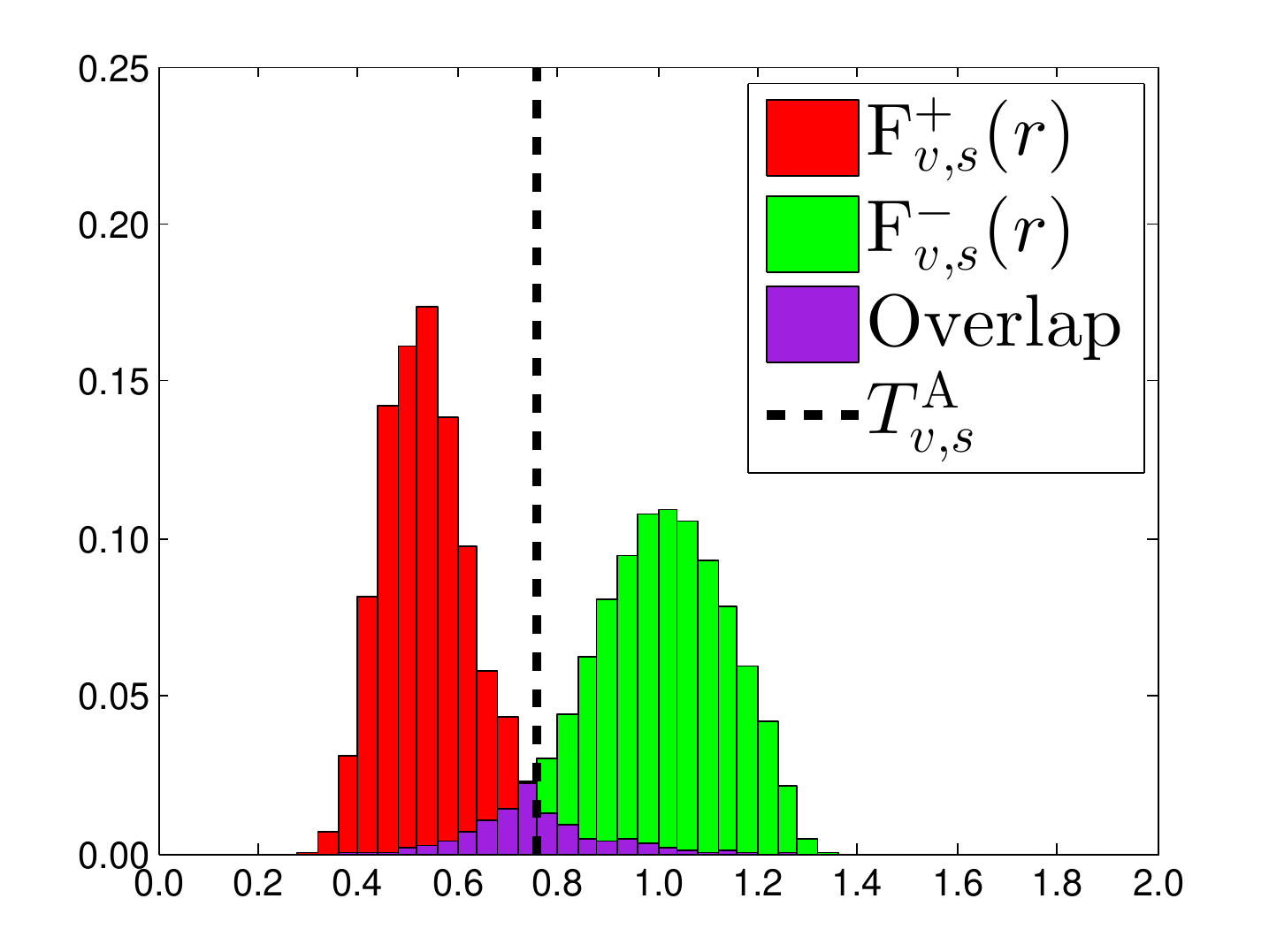}
    \includegraphics[width=1.2in]{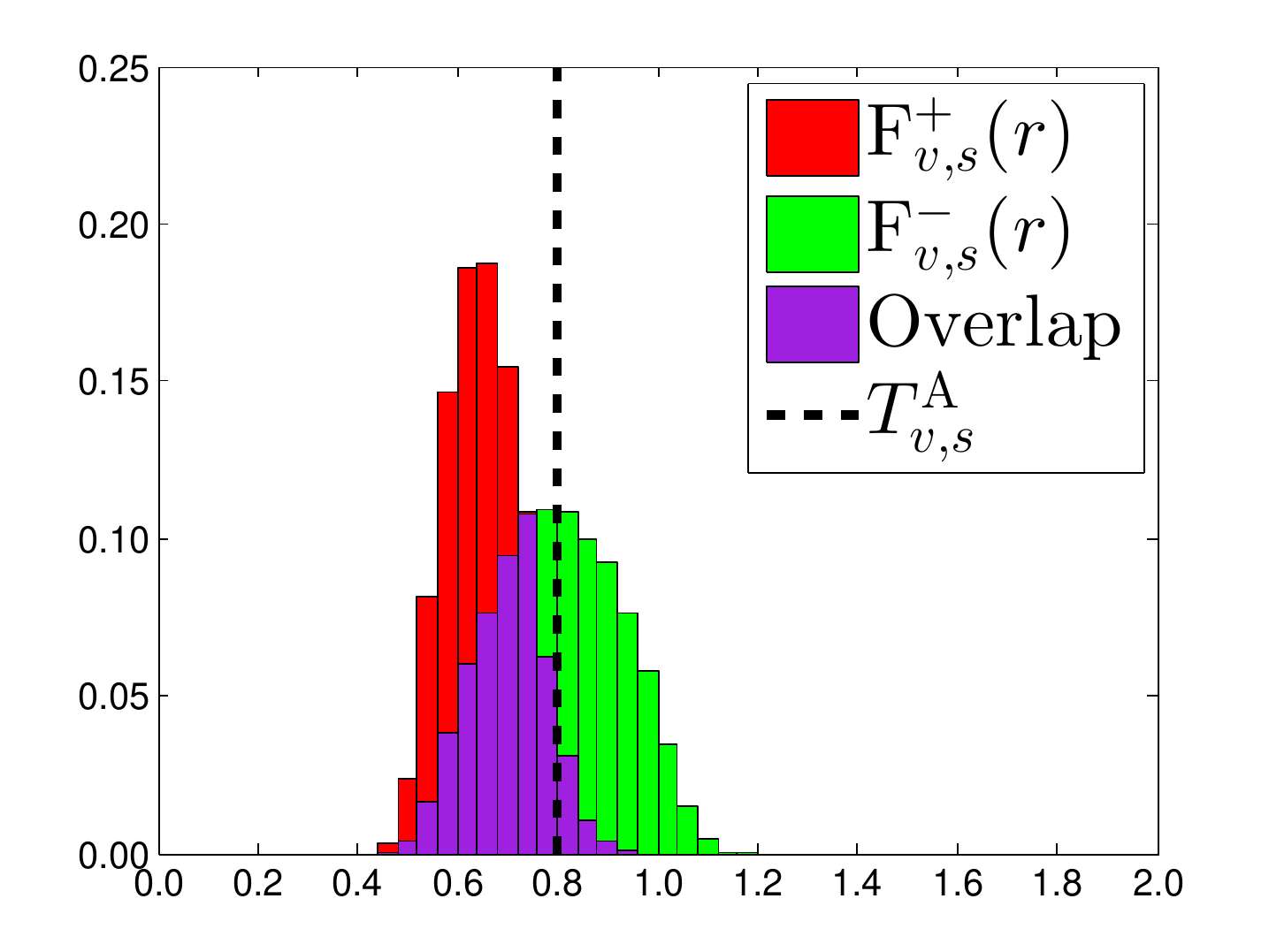}
    \includegraphics[width=1.2in]{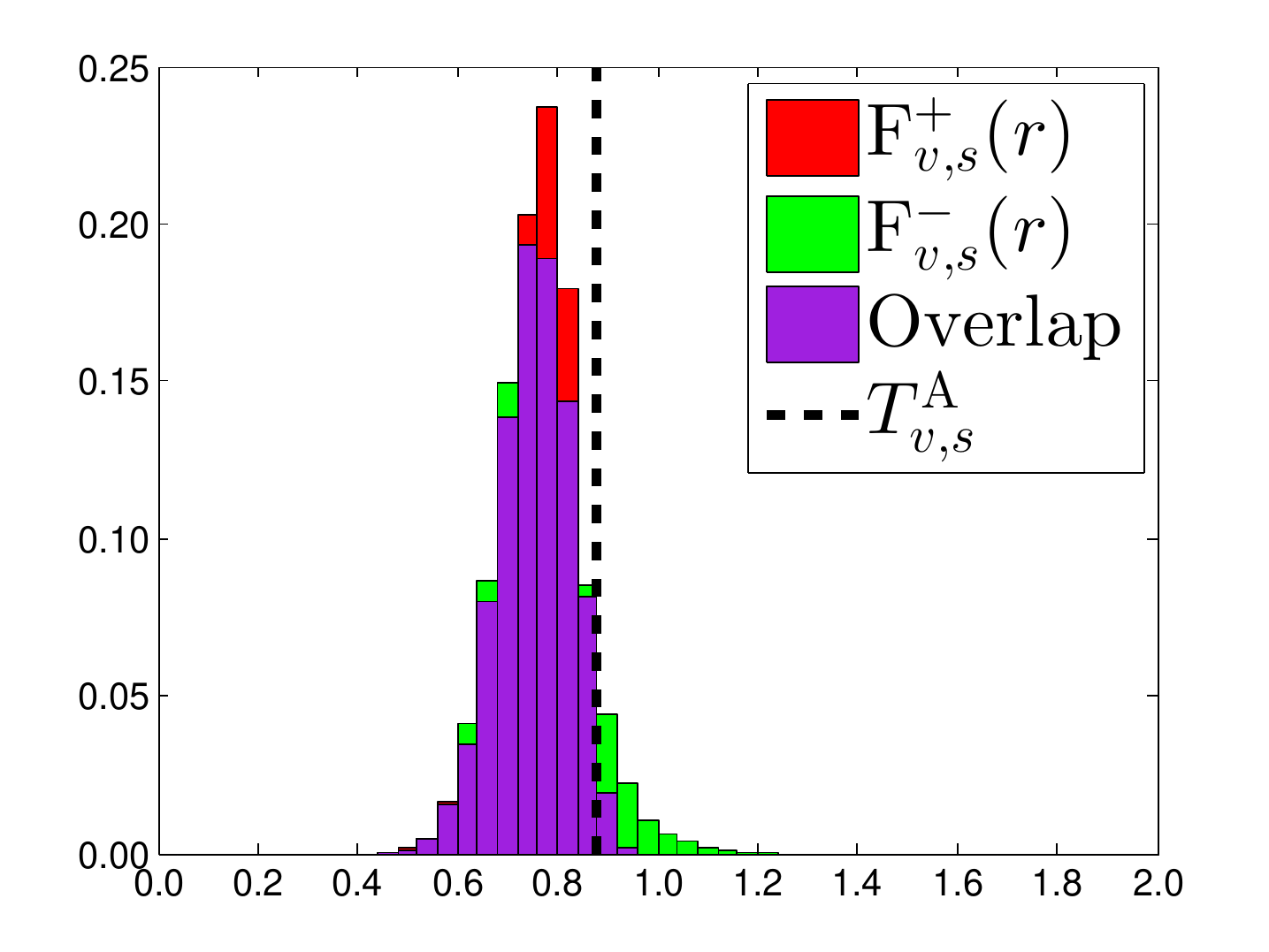}
    \includegraphics[width=1.2in]{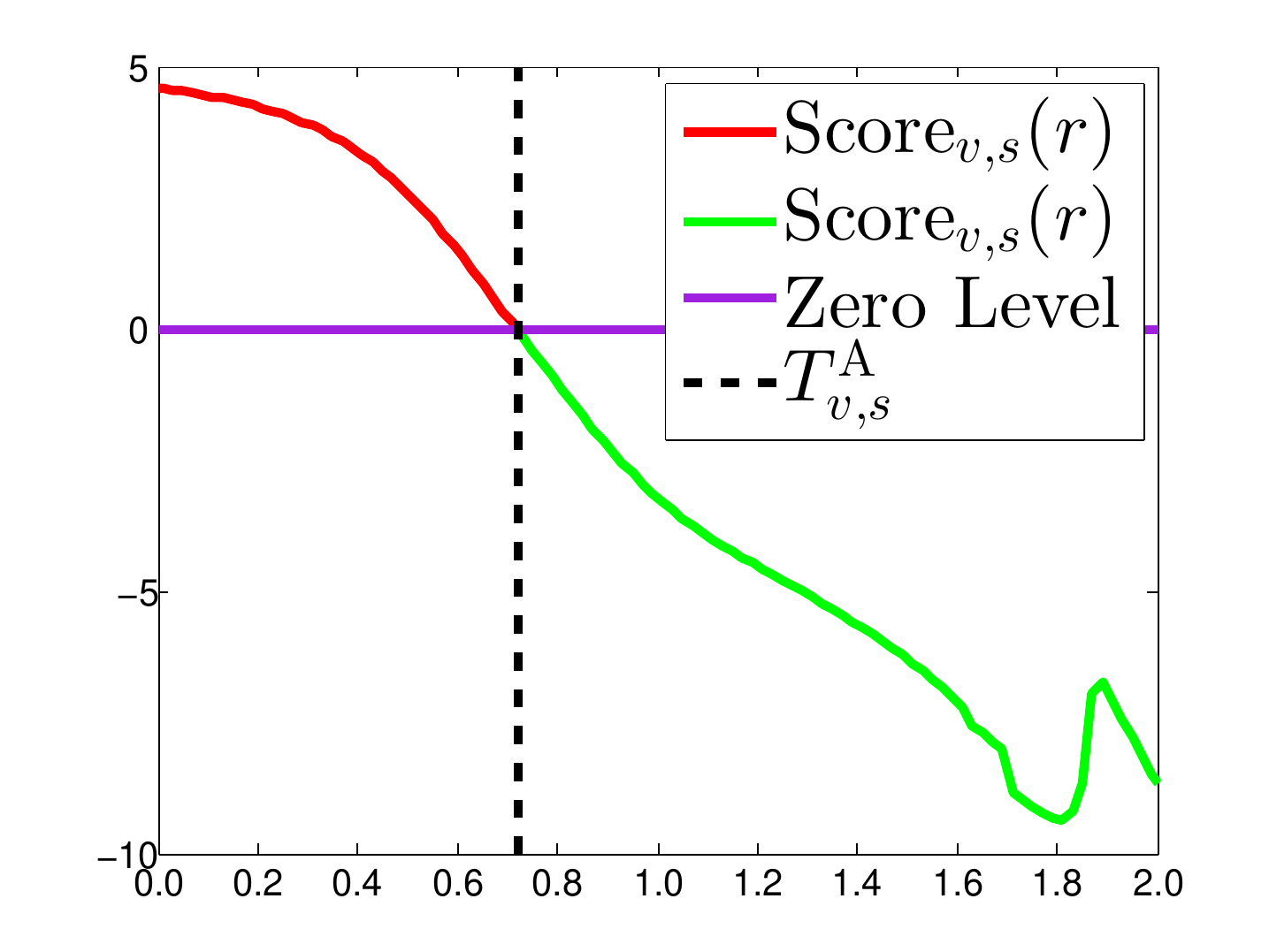}
    \includegraphics[width=1.2in]{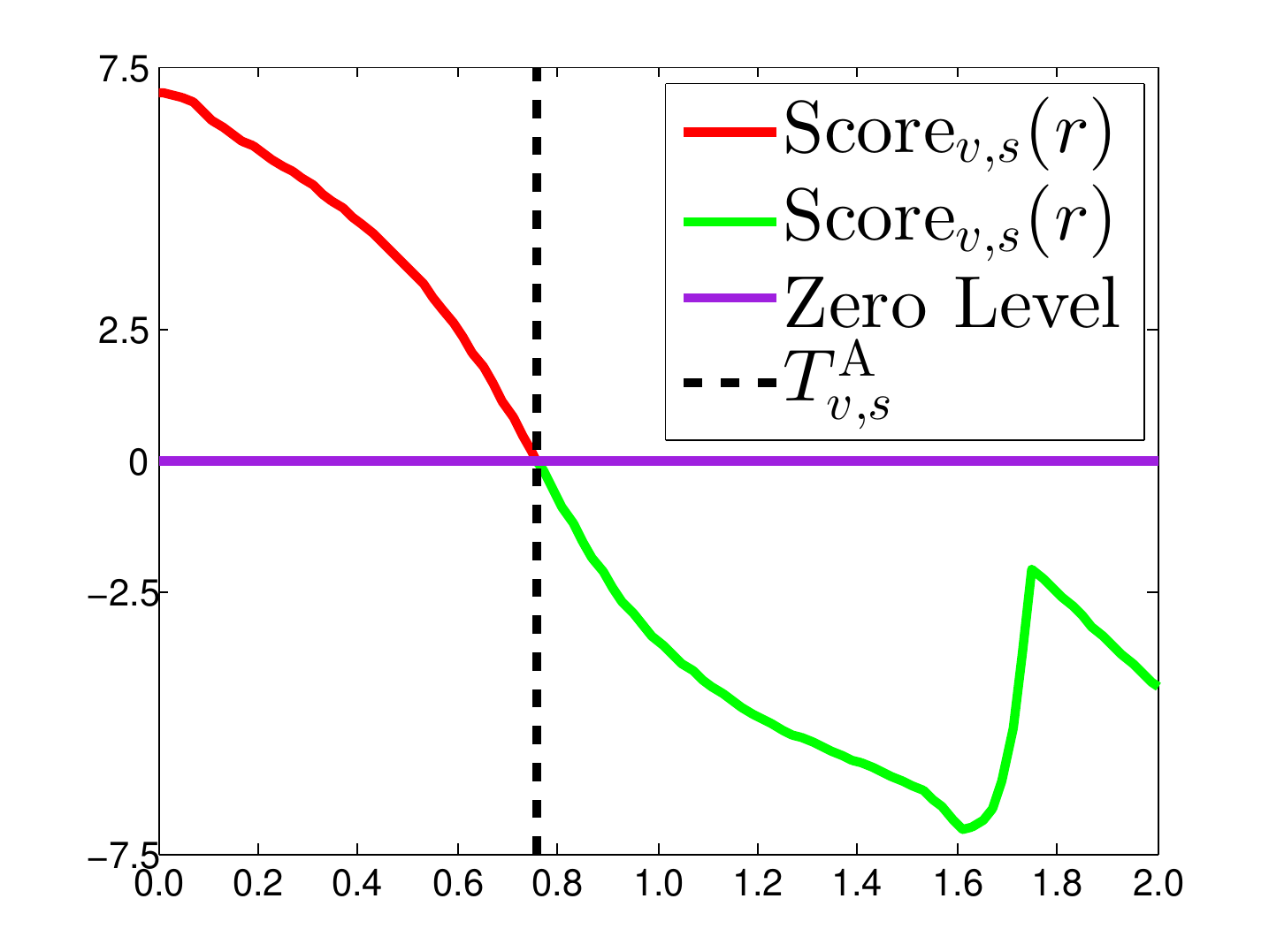}
    \includegraphics[width=1.2in]{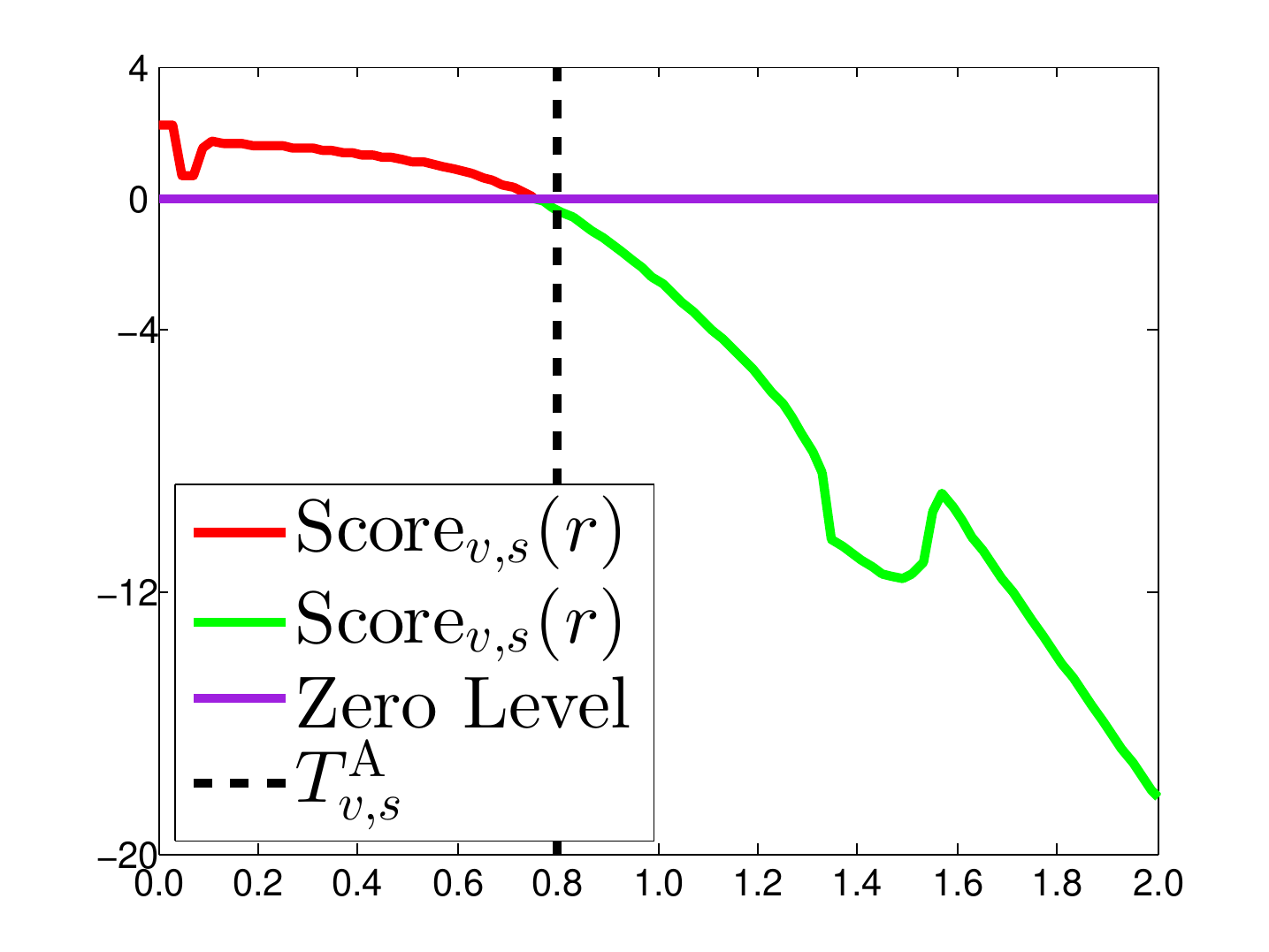}
    \includegraphics[width=1.2in]{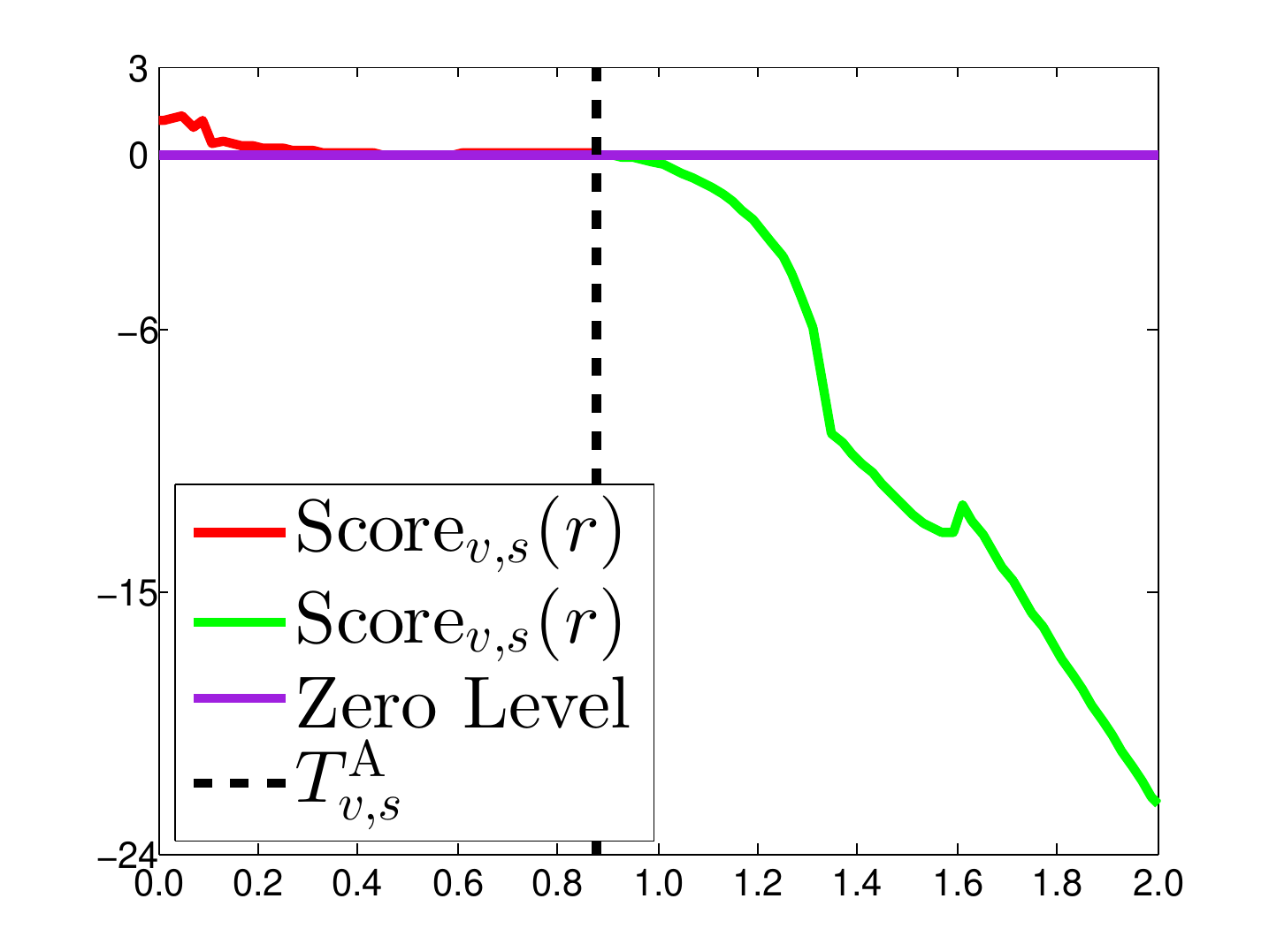}
\caption{
    The visual cues obtained in the training process (best viewed in color PDF).
    From left to right, we visualize the results of using $\mathrm{VC}_1$,
    $\mathrm{VC}_8$, $\mathrm{VC}_{60}$ and $\mathrm{VC}_{166}$ to support $\mathrm{SP}_1$.
    Of these, $\mathrm{VC}_1$ and $\mathrm{VC}_8$ are good supporters,
    $\mathrm{VC}_{60}$ is moderate, while $\mathrm{VC}_{166}$ is poor.
    Top row: the voting offset maps $\mathcal{H}_{v,s}$.
    Middle row: the target distributions $\mathrm{F}_{v,s}^+\!\left(r\right)$ (in red and purple)
    and the reference distributions $\mathrm{F}_{v,s}^-\!\left(r\right)$ (in green and purple).
    The overlap of these distributions reflects whether this VC is good at detecting this SP (smaller overlap is better).
    Bottom row: the score functions $\mathrm{Score}_{v,s}\!\left(r\right)$
    (red and green parts indicate positive and negative scores, respectively).
}
\label{Fig:Training}
\end{figure*}

For each $\mathrm{SP}_s$,
we select a positive training set $\mathcal{T}_s^+$ composing of those annotated positions $q$,
and a negative set $\mathcal{T}_s^-$ composing of a number of positions which are far away from any ground-truth positions.
For each $\mathrm{VC}_v$, we perform statistics on the positive and negative samples,
based on the previously defined neighborhoods $\mathcal{N}_{v,s}\!\left(q\right)$,
and compute the following conditional distributions:
\begin{equation}
\label{Eqn:ProbPositive}
{\mathrm{F}_{v,s}^+\!\left(r\right)}=
{\displaystyle{\frac{\mathrm{d}}{\mathrm{d}r}}\mathrm{Pr}\!\left[\displaystyle{{\min_{p\in\mathcal{N}_{v,s}\!\left(q\right)}}}
    \left\|\mathbf{f}\!\left(\mathbf{I}_p\right)-\mathbf{f}_v\right\|\leqslant r\mid q\in\mathcal{T}_s^+\right]},
\end{equation}
\begin{equation}
\label{Eqn:ProbNegative}
{\mathrm{F}_{v,s}^-\!\left(r\right)}=
{\displaystyle{\frac{\mathrm{d}}{\mathrm{d}r}}\mathrm{Pr}\!\left[\displaystyle{{\min_{p\in\mathcal{N}_{v,s}\!\left(q\right)}}}
    \left\|\mathbf{f}\!\left(\mathbf{I}_p\right)-\mathbf{f}_v\right\|\leqslant r\mid q\in\mathcal{T}_s^-\right]}.
\end{equation}
Here, the first distribution $\mathrm{F}_{v,s}^+\!\left(r\right)$ is the {\em target} distribution,
giving the activation pattern for $\mathrm{VC}_v$ if there is a semantic part $\mathrm{SP}_s$ nearby.
The intuition is that if $\left(\mathrm{VC}_v,\mathrm{SP}_s\right)$ is a good pair,
then the probability $\mathrm{F}_{v,s}^+\!\left(r\right)$ will be peaked close to ${r}={0}$,
({\em i.e.}, there will be some feature vectors within $\mathcal{N}_{v,s}\!\left(q\right)$ that cause $\mathrm{VC}_v$ to activate).
The second distribution, $\mathrm{F}_{v,s}^-\!\left(r\right)$ is the {\em reference} distribution
which specifies the response of the feature vector if the semantic part is not present.
This is needed~\cite{Konishi_1999_Fundamental}
to quantify the chance that we get a good match ({\em i.e.}, a small value of $r$) when the semantic part is not present.
In practice, we model the distribution using a histogram.
Some typical feature distributions are shown in Figure~\ref{Fig:Training}.
Note that the overlap between the target and reference distributions
largely reflects the quality of a $\left(\mathrm{VC}_v,\mathrm{SP}_s\right)$ pair.

With the probabilistic distributions,
we can find the set of supporting visual concepts ${\mathcal{V}_s}\subseteq{\mathcal{V}}$ for each semantic part $\mathrm{SP}_s$.
This is determined by first computing the threshold $T_{v,s}^\mathrm{A}$
that makes the false-negative rate ${\mathrm{FNR}_{v,s}}={5\%}$.
This threshold will be used in the testing phase to judge if $\mathrm{VC}_v$ fires at some grid position for $\mathrm{SP}_s$.
Note that we set a small FNR so that the positive samples are mostly preserved.
Although this may introduce some false positives, the voting model allows us to filter them out at a higher level.
Then, the top-$K$ visual concepts with the minimum false-positive rates $\mathrm{FPR}_{v,s}$ are selected to support $\mathrm{SP}_s$.
We fix ${K}={45}$, {\em i.e.}, ${\left|\mathcal{V}_s\right|}={45}$ for all ${s}={1,2,\ldots,\left|\mathcal{S}\right|}$.
We set a relatively large $K$ (in comparison to ${N}\approx{200}$),
so that when some of the supporting visual concepts are absent because of occlusion,
it is still possible to detect the semantic part via the present ones.

Not all supporting visual concepts are equally good.
We use the {\em log-likelihood ratio test}~\cite{Amit_2002_2D} to define a {\em score function}:
\begin{equation}
\label{Eqn:ScoreFunction}
{\mathrm{Score}_{v,s}\!\left(r\right)}=
    {\log\frac{\mathrm{F}_{v,s}^+\!\left(r\right)+\varepsilon}{\mathrm{F}_{v,s}^-\!\left(r\right)+\varepsilon}}.
\end{equation}
Here, ${\varepsilon}={10^{-7}}$ is a small floating point number to avoid invalid arithmetic operations.
The score function determines the evidence (either positive or negative) with respect to the feature distance,
{\em i.e.}, ${r}={\left\|\mathbf{f}\!\left(\mathbf{I}_p\right)-\mathbf{f}_v\right\|}$.
The visualization of these scores are shown in Figure~\ref{Fig:Training}.

In summary, the following information is learned in the training process.
For each semantic part $\mathrm{SP}_s$, a set of supporting visual concepts is learned.
For each $\left(\mathrm{VC}_v,\mathrm{SP}_s\right)$ pair,
we obtain the voting offset map $\mathcal{H}_{v,s}$, the activation threshold $T_{v,s}^\mathrm{A}$,
and the score function $\mathrm{Score}_{v,s}\!\left(r\right)$.
These will be used in the testing stage.

\subsection{The Testing Phase}
\label{Algorithm:Testing}

Now, given a testing image $\mathbf{I}$, our goal is to detect all semantic parts on it.
As in the training phase, each semantic part $\mathrm{SP}_s$ is processed individually.
Recall that we have obtained the set of supporting visual concepts $\mathcal{V}_s$.
For each supporting ${\mathrm{VC}_v}$, an individual voting process is performed on the entire image.

The voting process starts with extracting CNN features on the {\em pool-4} layer.
Let $\mathbf{f}\!\left(\mathbf{I}_p\right)$ be a feature vector at a position $p \in \mathcal{L}_4$.
We test if this feature vector activates ${\mathrm{VC}_v}$
by checking if the feature distance $\left\|\mathbf{f}\!\left(\mathbf{I}_p\right)-\mathbf{f}_v\right\|$
is smaller than the activation threshold $T_{v,s}^\mathrm{A}$ obtained in the training process.
If it is activated, we make use of the score function $\mathrm{Score}_{v,s}\!\left(r\right)$,
and substitute with ${r_v\!\left(\mathbf{I}_p\right)}={\left\|\mathbf{f}\!\left(\mathbf{I}_p\right)-\mathbf{f}_v\right\|}$.
This term, $\mathrm{Score}_{v,s}\!\left(r_v\!\left(\mathbf{I}_p\right)\right)$,
or $\mathrm{Score}_{v,s}\!\left(\mathbf{I}_p\right)$ for short,
is the evidence that $\mathbf{f}\!\left(\mathbf{I}_p\right)$ votes for $\mathrm{SP}_s$.
It is added to various positions at the {\em pool-4} layer determined by the offset map $\mathcal{H}_{v,s}$.
Besides, recall that $\mathcal{H}_{v,s}$ is a set of offset vectors,
each of them, denoted as $\Delta p$, is equipped with a frequency $\mathrm{Fr}\!\left(\Delta p\right)$.
The final score which is added to the position $p+\Delta p$ is computed as:
\begin{equation}
\label{Eqn:Voting}
{\mathrm{Vote}_{v,s}\!\left(p+\Delta p\right)}=
    {\left(1-\beta\right)\mathrm{Score}_{v,s}\!\left(\mathbf{I}_p\right)+\beta\log\frac{\mathrm{Fr}\!\left(\Delta p\right)}{U}}.
\end{equation}
The first term ensures that there is high evidence of $\mathrm{VC}_v$ firing,
and the second term acts as the spatial penalty ensuring that this $\mathrm{VC}_v$ fires on the right position
specified by the offset map $\mathcal{H}_{v,s}$.
Here we set ${\beta}={0.7}$,
and define ${\log\frac{\mathrm{Fr}\!\left(\Delta p\right)}{U}}={-\infty}$
when ${\frac{\mathrm{Fr}\!\left(\Delta p\right)}{U}}={0}$.
$U$ is a constant which is the average frequency over the entire offset map $\mathcal{H}_{v,s}$.


After all activated positions of $\mathrm{VC}_v$ are considered,
we combine the voting results by preserving the maximal response at each position $p$.
If the maximal response is negative at one position, it is set to $0$ to avoid introducing negative cues,
{\em i.e.}, that a visual concept is allowed to support a semantic part, but not allowed to inhibit it.
The final score for detecting $\mathrm{SP}_s$ involves summing up the voting results of all supporting visual concepts:
\begin{equation}
\label{Eqn:FinalScore}
{\mathrm{Score}_s\!\left(p\right)}=
    {{\sum_{\mathrm{VC}_v\in\mathcal{V}_s}}\max\left\{0,\mathrm{Vote}_{v,s}\!\left(p\right)\right\}}.
\end{equation}
This score map is computed at the {\em pool-4} layer.
It is then resized to the original image size using 2D spline interpolation.

\subsubsection{The Multi-Scale Testing Strategy}
\label{Algorithm:Testing:MultiScale}

Note that the training process is performed with the object bounding boxes provided,
which limits our algorithm's ability of detecting a semantic part at a different scale from the training case.
To deal with this issue, we design a multi-scale voting scheme.

This scheme is only used in testing, {\em i.e.}, no extra training is required.
Given a testing image, we resize it to $10$ different scales,
with the short edge containing $224$, $272$, $320$, $400$, $480$, $560$, $640$, $752$, $864$ and $976$ pixels, respectively.
A larger number of scales may lead to better detection results but the computational cost becomes more expensive.
Then, we run the testing process at each scale and get $10$ score maps. For each $\mathrm{SP}_s$,
we find the highest detection score among all score maps at different scales.
We denote the scale producing the highest score as $\mathrm{Sc}_s$.
This provides evidence for the proper scale of the image.
The final scale of the image is obtained by averaging all such evidences,
{\em i.e.}, computing ${\mathrm{Sc}^\star}={\frac{1}{\mathcal{S}}{\sum_{s\in\mathcal{S}}}\mathrm{Sc}_s}$.
We use average rather than max operation in order to take multi-scale information and improve the robustness of our approach.
Finally, we resize the image based on the predicted scale $\mathrm{Sc}^\star$,
and run the detection process again.
We will show in Section~\ref{Experiments:NoOcclusion} that this simple method works well.

\section{Experiments}
\label{Experiments}

\newcommand{\colwidth}{0.63cm}
\begin{table}
\centering
\begin{tabular}{|l||R{\colwidth}|R{\colwidth}|R{\colwidth}|R{\colwidth}||
                    R{\colwidth}|R{\colwidth}|R{\colwidth}||
                    R{\colwidth}|R{\colwidth}|R{\colwidth}|}
\hline
\multirow{2}{*}{} & \multicolumn{4}{c||}{Natural Detection}
                  & \multicolumn{3}{c||}{Oracle Detection}
                  & \multicolumn{3}{c| }{Scale Pred. Loss}                               \\
\cline{2-11}
{Object}          & \multicolumn{1}{c|}{\bf VC}         & \multicolumn{1}{c|}{\bf SV}
                  & \multicolumn{1}{c|}{\bf FR}         & \multicolumn{1}{c||}{\bf VT}
                  & \multicolumn{1}{c|}{\bf VC}
                  & \multicolumn{1}{c|}{\bf SV}         & \multicolumn{1}{c||}{\bf VT}
                  & \multicolumn{1}{c|}{\bf VC}
                  & \multicolumn{1}{c|}{\bf SV}         & \multicolumn{1}{c|}{\bf VT}         \\
\hline\hline
{\em airplane}    & $10.1 $          & $18.2 $          & $\mathbf{44.9 }$ & $30.6 $
                  & $18.5 $          & $25.9 $          & $\mathbf{41.1 }$
                  & $0.27 $          & $0.22 $          & $\mathbf{0.21 }$                    \\
\hline
{\em bicycle}     & $48.0 $          & $58.1 $          & $\mathbf{78.4} $          & $77.8 $
                  & $61.8 $          & $73.8 $          & $\mathbf{81.6 }$
                  & $0.20 $          & $0.20 $          & $\mathbf{0.13 }$                    \\
\hline
{\em bus}         & $ 6.8 $          & $26.0 $          & $\mathbf{65.3} $  & $58.1 $
                  & $27.8 $          & $39.6 $          & $\mathbf{60.3 }$
                  & $0.32 $          & $0.21 $          & $\mathbf{0.14 }$                    \\
\hline
{\em car}         & $18.4 $          & $27.4 $          & $\mathbf{68.4 }$ & $63.4 $
                  & $28.1 $          & $37.9 $          & $\mathbf{65.8 }$
                  & $0.23 $          & $0.21 $          & $\mathbf{0.11 }$                    \\
\hline
{\em motorbike}   & $10.0 $          & $18.6 $          & $47.7 $          & $\mathbf{53.4 }$
                  & $34.0 $          & $43.8 $          & $\mathbf{58.7 }$
                  & $0.35 $          & $0.31 $          & $\mathbf{0.18 }$                    \\
\hline
{\em train}       & $ 1.7 $          & $ 7.2 $          & $\mathbf{42.9 }$ & $35.5 $
                  & $13.6 $          & $21.1 $          & $\mathbf{51.4 }$
                  & $0.40 $          & $0.29 $          & $\mathbf{0.22 }$                    \\
\hline
{\bf mean}        & $15.8 $          & $25.9 $          & $\mathbf{58.0} $ & $53.1 $
                  & $30.6 $          & $40.4 $          & $\mathbf{59.8 }$
                  & $0.30 $          & $0.24 $          & $\mathbf{0.17 }$                    \\
\hline
\end{tabular}
\caption{
    Detection accuracy (mean AP, $\%$) and scale prediction loss without occlusion.
}
\label{Tab:NoOcclusion}
\end{table}

\subsection{Settings}
\label{Experiments:Settings}

\noindent {\bf Dataset.}
We use VehicleSemanticPart dataset~\cite{Wang_2015_Unsupervised} for training.
It contains non-occluded images with dense part labeling of $6$ objects,
{\em i.e.}, {\em airplane}, {\em bicycle}, {\em bus}, {\em car}, {\em motorbike} and {\em train}.
Some typical semantic parts are shown in Figure~\ref{Fig:Goal}.
For fair comparison, all the algorithms are trained on the bounding-box images without occlusions.
As for the test set, we randomly superimpose two, three or four irrelevant objects (named {\em occluders}) onto the target object.
{\bf We use such synthesized images
because they are easy to generate and the actual position of the occluded parts can be accurately annotated.}
We also control the occlusion ratio by computing the fraction of occluded pixels on the target object.

\noindent {\bf Criterion.}
We evaluate our algorithm in two cases, {\em i.e.}, whether the target object is partially occluded.
We follow the most popular criteria~\cite{Everingham_2010_PASCAL},
where a detected semantic part is true-positive if it matches a ground-truth annotation,
{\em i.e.}, the Intersection-over-Union (IoU) ratio between two boxes is not smaller than $0.5$.
Duplicate detection is counted as false-positive.

\noindent {\bf Baselines.}
In all experiments, our algorithm (denoted as {\bf VT}) is compared to three baseline approaches,
{\em i.e.}, single visual concept detection~\cite{Wang_2015_Unsupervised}, Faster-RCNN~\cite{Ren_2015_Faster}, and SVM+LLC.
The first one (denoted as {\bf VC}) follows the exact implementation in~\cite{Wang_2015_Unsupervised}.
The second one (denoted as {\bf FR}) involves re-training a Faster-RCNN~\cite{Ren_2015_Faster} model for each of the six classes,
{\em i.e.}, each semantic part is considered as an ``object category''.
In the third baseline (denoted as {\bf SV}), we follows the standard Bag-of-Visual-Words (BoVW) model,
and train a binary SVM classifier for detecting each semantic part.
We first encode each CNN feature $\mathbf{f}\!\left(\mathbf{I}_p\right)$
into a $\left|\mathcal{V}\right|$-dimensional vector $\mathbf{v}_p$
using Locality-sensitive Linear Coding (LLC)~\cite{Wang_2010_Locality}.
The number of bases in LLC is set to be $45$, {\em i.e.}, the number of supporting visual concepts for each semantic part.
Then, following the flowchart in Section~\ref{Algorithm:Training:SpatialRelationship},
we select a positive set $\mathcal{T}_s^+$ and a negative set $\mathcal{T}_s^-$,
and compute the feature vector $\mathbf{v}_p$ at each position and train a SVM classifier.
At the testing stage, we compute the feature vector at each position, feed it to the SVM,
and finally compose the detection score map with the confidence scores provided by the binary SVM.
We do not consider some part-based models~\cite{Fidler_2007_Towards}\cite{Felzenszwalb_2010_Object},
as they are not based on deep network features, and thus do not produce state-of-the-art detection accuracy.

\subsection{Semantic Part Detection without Occlusion}
\label{Experiments:NoOcclusion}

We first assume that the target object is not occluded by any irrelevant objects.
Results of our algorithm and its competitors are summarized in Table~\ref{Tab:NoOcclusion}.
Our voting algorithm achieves comparable detection accuracy to Faster-RCNN~\cite{Ren_2015_Faster},
one of the state-of-the-art object detectors.
Note that Faster-RCNN~\cite{Ren_2015_Faster} depends on some discriminative information,
while our voting algorithm relies only on integrating visual cues from visual concepts.
As we will see later, our algorithm works better than Faster-RCNN~\cite{Ren_2015_Faster} on the occlusion cases.
Since~\cite{Wang_2015_Unsupervised} merely uses single visual concepts for detection,
it produces significantly lower accuracy than our approach.
SVM+LLC uses a strong classifier, but produces unsatisfying performance due to the lack of considering context information.

\renewcommand{\colwidth}{0.77cm}
\begin{table*}
\centering
\begin{tabular}{|l||R{\colwidth}|R{\colwidth}|R{\colwidth}||
                    R{\colwidth}|R{\colwidth}|R{\colwidth}||
                    R{\colwidth}|R{\colwidth}|R{\colwidth}|}
\hline
\multirow{2}{*}{} & \multicolumn{3}{c||}{$2$ Occ's, ${0.2}\leqslant{r}<{0.4}$}
                  & \multicolumn{3}{c||}{$3$ Occ's, ${0.4}\leqslant{r}<{0.6}$}
                  & \multicolumn{3}{c| }{$4$ Occ's, ${0.6}\leqslant{r}<{0.8}$}            \\
\cline{2-10}
{Object}          & \multicolumn{1}{c|}{\bf SV}
                  & \multicolumn{1}{c|}{\bf FR}         & \multicolumn{1}{c||}{\bf VT}
                  & \multicolumn{1}{c|}{\bf SV}
                  & \multicolumn{1}{c|}{\bf FR}         & \multicolumn{1}{c||}{\bf VT}
                  & \multicolumn{1}{c|}{\bf SV}
                  & \multicolumn{1}{c|}{\bf FR}         & \multicolumn{1}{c|}{\bf VT}         \\
\hline\hline
{\em airplane}    & $12.0 $          & $\mathbf{26.8 }$ & $23.2 $
                  & $ 9.7 $          & $\mathbf{20.5 }$ & $19.3 $
                  & $ 7.5 $          & $\mathbf{15.8 }$ & $15.1 $          \\
\hline
{\em bicycle}     & $44.6 $          & $65.7 $          & $\mathbf{71.7 }$
                  & $33.7 $          & $54.2 $          & $\mathbf{66.3 }$
                  & $15.6 $          & $37.7 $          & $\mathbf{54.3 }$ \\
\hline
{\em bus}         & $12.3 $          & $\mathbf{41.3 }$ & $31.3 $
                  & $ 7.3 $          & $\mathbf{32.5 }$ & $19.3 $
                  & $ 3.6 $          & $\mathbf{21.4 }$ & $ 9.5 $          \\
\hline
{\em car}         & $13.4 $          & $\mathbf{35.9 }$ & $\mathbf{35.9 }$
                  & $ 7.7 $          & $22.0 $          & $\mathbf{23.6 }$
                  & $ 4.5 $          & $\mathbf{14.2 }$ & $13.8 $          \\
\hline
{\em motorbike}   & $11.4 $          & $35.9 $          & $\mathbf{44.1 }$
                  & $ 7.9 $          & $28.8 $          & $\mathbf{34.7 }$
                  & $ 5.0 $          & $19.1 $          & $\mathbf{24.1 }$ \\
\hline
{\em train}       & $ 4.6 $          & $20.0 $          & $\mathbf{21.7 }$
                  & $ 3.4 $          & $\mathbf{11.1 }$ & $ 8.4 $
                  & $ 2.0 $          & $\mathbf{ 7.2 }$ & $ 3.7 $          \\
\hline
{\bf mean}        & $16.4 $          & $37.6 $          & $\mathbf{38.0 }$
                  & $11.6 $          & $28.2 $          & $\mathbf{28.6 }$
                  & $ 6.4 $          & $19.2 $          & $\mathbf{20.1 }$ \\
\hline
\end{tabular}
\caption{
    Detection accuracy (mean AP, $\%$) when the object is partially occluded. Three levels of occlusion are considered.
}
\label{Tab:Occlusion}
\end{table*}

\noindent {\bf Scale Prediction.}
Since all methods are trained on cropped bounding-box images,
they lack the ability of detecting semantic parts at a different scale from the training case.
For Faster-RCNN~\cite{Ren_2015_Faster}, we construct an image pyramid,
where the short side of an image will be resized to 5 scales,
{\em i.e.}, $\{600, 688, 800, 976, 1200\}$, to fuse detection results at test time.
However, we cap the longest side at $3000$ pixels to avoid exceeding GPU memory.
For other methods, we applied the same scale prediction algorithm,
which is described in Section~\ref{Algorithm:Testing:MultiScale}.
To illustrate the importance of scale prediction algorithm, we present an {\em oracle} detection option,
which resizes each image according to the ground-truth bounding box size,
{\em i.e.}, after rescaling, the short edge of the target object becomes $224$, as in the training case.
As we can see in Table~\ref{Tab:NoOcclusion}, this improves the performance of all three competitors significantly.

To analyze the accuracy of scale prediction, we perform a diagnostic experiment.
For each testing image, we use the ground-truth bounding box to compute the {\em actual size} it should be rescaled into.
For example, if a {\em car} occupies a $150\times100$ region in a $200\times250$ image,
given that $150\times100$ is rescaled to $336\times224$ (as in training), the full image should be rescaled to $448\times560$.
If the short edge is $a$ for the actual size and $b$ in scale prediction,
then the loss in rescaling is computed as $\ln\!\left(\max\left\{a,b\right\}/\min\left\{a,b\right\}\right)$.
A perfect prediction has a loss of $0$.
Results are summarized in Table~\ref{Tab:NoOcclusion}.
The voting algorithm predicts the object scale more accurately, leading to a higher overall detection accuracy,
and the least accuracy drop without using the oracle information.

\noindent {\bf Diagnosis.}
We diagnose the voting algorithm by analyzing the contribution of some modules.
Two options are investigated with oracle information.
First, changing the number of supporting visual concepts.
We decrease $\left|\mathcal{V}_s\right|$ from $45$ to $30$, $20$ and $10$,
and observe $1.0\%$, $3.3\%$ and $7.8\%$ mean accuracy drop, respectively.
On the other hand, increasing it from $45$ to $60$ does not impact the performance much ($0.4\%$ improvement).
Similar phenomena are also observed when occlusion is present.
Therefore, we conclude that the voting method requires a sufficient number of supporting visual concepts,
but using too many of them may introduce redundant information which increases the computational overhead.
Second, We consider another smaller neighborhood threshold $\gamma_{\mathrm{th}}=56$,
which leads to a spatial offset map of size $7\times7$.
This causes $4.5\%$ accuracy drop,
arguably caused by the lack of long-distance visual cues which is useful especially when occlusion is present.

\subsection{Semantic Part Detection under Occlusion}
\label{Experiments:Occlusion}

Next, we investigate the case that the target object is partially occluded.
We construct $3$ datasets with different occluder numbers and occlusion ratios.
We still apply the multi-scale detection described in Section~\ref{Algorithm:Testing:MultiScale}.
Note that all models are trained on the non-occlusion dataset,
and we will report the results with occluded training objects in the future.

Results are shown in Table~\ref{Tab:Occlusion}.
To save space,
we ignore the performance of {\bf VC}~\cite{Wang_2015_Unsupervised} since it is the weakest one among all baseline methods.
As occlusion becomes heavier, we observe significant accuracy drop.
Note that all these methods are trained on the non-occlusion dataset.
Since our voting algorithm has the ability of inferring the occluded parts via its contexts, the accuracy drop is the smallest.
Consequently, its advantage in detection accuracy becomes more significant over other competitors.

\section{Conclusions and Future Work}
\label{Conclusions}

We address the task of detecting semantic parts under occlusions.
We design a novel framework which involves modeling spatial relationship, finding supporting visual concepts,
performing log-likelihood ratio tests, and summarizing visual cues with a voting process.
Experiments verify that our algorithm works better than previous work when the target object is partially occluded.
Our algorithm also enjoys the advantage of being explainable,
which is difficult to achieve in the state-of-the-art holistic object detectors.

In the future, we will extend our approach to detect the entire object under occlusion.
In our preliminary experiments,
we have already observed that proposal-based methods such as Faster-RCNN are not robust to heavy occlusion on the entire object.
This can be dealt with in two different ideas,
{\em i.e.}, merging the detected semantic parts in a bottom-up manner,
or using a pre-defined template to organize the detected semantic parts.
We will also try to construct a dataset with real occluded images, and a dataset with more object classes.

\vspace{0.2cm}
\noindent
{\bf Acknowledgements.}
This work is supported by the Intelligence Advanced Research Projects Activity (IARPA) via DoI/IBC contract number D16PC00007,
and also by the NSF STC award CCF-1231216.
We thank Dr. Vittal Premachandran for his enormous help in this project.
We also thank Dr. Wei Shen, Dr. Ehsan Jahangiri, Dr. Yan Wang,
Weichao Qiu, Chenxi Liu, Zhuotun Zhu, Siyuan Qiao and Yuyin Zhou for instructive discussions.

\bibliography{egbib}
\end{document}